\documentclass[a4paper,conference]{IEEEtran}
\makeatletter

\usepackage{graphics}
\usepackage{amsmath}
\usepackage{caption}
\usepackage{amssymb}
\usepackage{times}
\usepackage{epsfig}
\DeclareMathOperator*{\argmin}{arg\,min}
\usepackage{multirow}
\usepackage{arydshln}
\usepackage{color}
\usepackage{comment}
\usepackage{booktabs}
\usepackage{dirtytalk}
\usepackage{amsfonts}

\makeatletter
\newcommand*{\rom}[1]{\expandafter\@slowromancap\romannumeral #1@}
\makeatother

\usepackage{cite}
\usepackage{algorithm}
\usepackage{algorithmicx}
\usepackage{algpseudocode}

\usepackage{array}

\usepackage{booktabs}

\begin{document}
%
\title{Self-Supervised Representation Learning for Visual Anomaly Detection}

\author{\IEEEauthorblockN{Rabia Ali}
\IEEEauthorblockA{KAIST\\
}
\and
\IEEEauthorblockN{Muhammad Umar Karim Khan}
\IEEEauthorblockA{KAIST\\
}
\and
\IEEEauthorblockN{Chong Min Kyung}
\IEEEauthorblockA{KAIST\\
}}


%


\maketitle

\begin{abstract}
   Self-supervised learning allows for better utilization of unlabelled data. The feature representation obtained by self-supervision can be used in downstream tasks such as classification, object detection, segmentation, and anomaly detection. While classification, object detection, and segmentation have been investigated with self-supervised learning, anomaly detection needs more attention. We consider the problem of anomaly detection in images and videos, and present a new visual anomaly detection technique for videos. Numerous seminal and state-of-the-art self-supervised methods are evaluated for anomaly detection on a variety of image datasets. The best performing image-based self-supervised representation learning method is then used for video anomaly detection to see the importance of spatial features in visual anomaly detection in videos. We also propose a simple self-supervision approach for learning temporal coherence across video frames without the use of any optical flow information. At its core, our method identifies the frame indices of a jumbled video sequence allowing it to learn the spatiotemporal features of the video. This intuitive approach shows superior performance of visual anomaly detection compared to numerous methods for images and videos on UCF101 and ILSVRC2015 video datasets.
\end{abstract}

\section{Introduction}

\begin{figure}[t]
\begin{center}
\includegraphics[height=3.8cm]{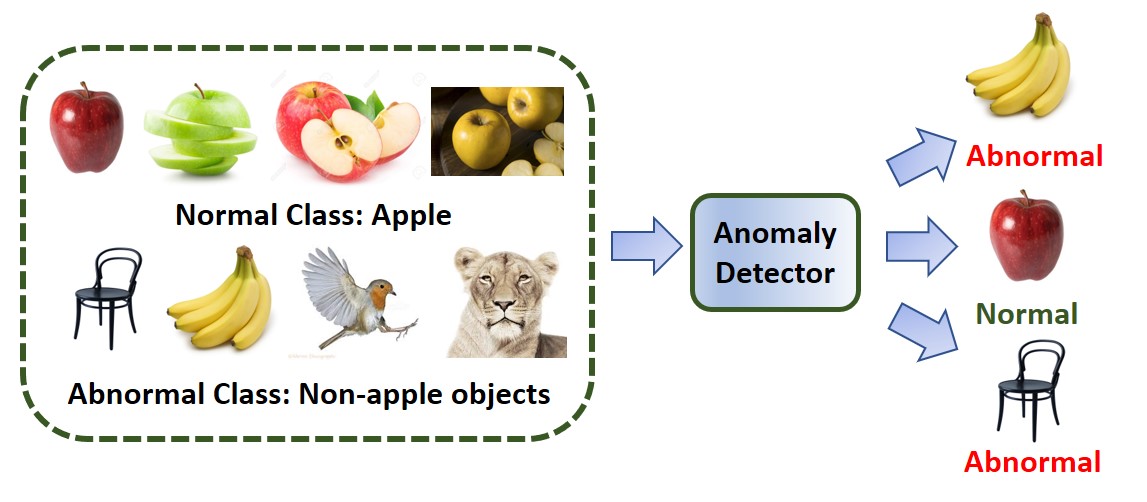}
\end{center}
\caption{In one-class classification (anomaly detection), the detector is expected to identify out-of-distribution (abnormal) instances. Here, given a set of normal apple instances, the detector learns to detect abnormal instances of non-apple objects}
\label{fig:example1}
\end{figure}

Anomaly detection, also termed as one-class classification, is a classic problem \cite{chalapathy2018anomaly-18,gong2019memorizing-16,schlegl2017unsupervised-17}. One-class classifiers are capable of identifying out-of-distribution (abnormal) instances by learning from the instances of the normal (in-distribution) class as shown in the  Fig.~\ref{fig:example1}. We address the problem of anomaly detection for images and videos, which is useful in applications such as visual quality inspection in manufacturing \cite{haselmann2018anomaly-64}, surveillance \cite{shashikar2017traffic-66,sultani2018real-65}, biomedical applications \cite{taboada2009anomaly-67,wei2018anomaly-68}, self-driving cars \cite{creusot2015real-69}, and robotics \cite{chakravarty2007anomaly-70,munawar2017spatio-71}. One-class classification is more general compared to binary classification because all unseen instances are treated as anomalies.

Anomalies in images and videos are generally defined as objects or events that are unusual and indicate an irregular behaviour. Manually detecting these rare objects in images and unexpected events in videos is a very tiresome task. Automating this cumbersome job, e.g., by self-supervised representation learning can ease the task of detecting: faulty materials in manufacturing, malignant tumor or a nodule from medical images including mammogram, CT or PET images, and anomalous events such as traffic accidents, crimes or illegal activities in video surveillance. 

Deep Neural Networks (DNNs) learn different levels of visual features in images and videos resulting in remarkable performance in classification \cite{cirecsan2012multi:2,cirecsan2011high:1,krizhevsky2012imagenet:3}, object detection \cite{girshick2015fast:5,girshick2014rich:4,ren2015faster:6}, semantic segmentation \cite{chen2017deeplab:8,long2015fully:7,zhao2017pyramid:9}, and anomaly detection \cite{chalapathy2019deep:12,kiran2018overview:11,kwon2017survey:10}. However, the near-human or superhuman performances achieved by deep learning algorithms generally requires annotation, which is both time-consuming and expensive. To reduce or avoid the cumbersome job of data labelling, researchers have for long focused on methods that require minimum level of supervision. These efforts led to advancements in domain adaptation, transfer learning, meta learning, continual learning, semi-supervised learning, weakly-supervised learning, and unsupervised learning.

We focus on self-supervised learning, a promising subclass of unsupervised learning, which provides an opportunity for better utilizing unlabeled data by setting the learning objectives to learn from the internal cues. In self-supervised learning a pretext task such as context prediction \cite{doersch2015unsupervised:13}, colorization \cite{zhang2016colorful-14}, predicting image rotations \cite{gidaris2018unsupervised-30}, etc. can be formulated by using only the unlabeled data. The network while solving these pretext tasks learns a useful feature representation which can be transferred to different downstream tasks of interest, such as classification, object detection, segmentation, and anomaly detection. 

In the context of deep learning, different schemes have been proposed for unsupervised anomaly detection, which includes using autoencoders \cite{an2015variational-21,hasan2016learning-48,xia2015learning-20,zong2018deep-19}, Generative Adversarial Networks (GANs) \cite{li2018anomaly-23}, and low-density rejection \cite{el2007optimal-39} and single-class SVM \cite{scholkopf2000support-40,tax2004support-41} over the low-dimensional embeddings. However, with the attempt to shift from supervised learning to unsupervised learning, anomaly detection using self-supervised learning is somewhat less explored. Anomaly detection using self-supervised visual representation learning aims to train a model such that it better learns the features of in-distribution (normal) examples. The learned features should not only focus on the low-level object characteristics like the color, texture, etc. but also on the high-level characteristics such as the object parts, shapes, position, and orientation. By learning these features from normal instances, the network can detect out-of-distribution samples.

\begin{figure*}[t]
\begin{center}
\includegraphics[height=7.0cm]{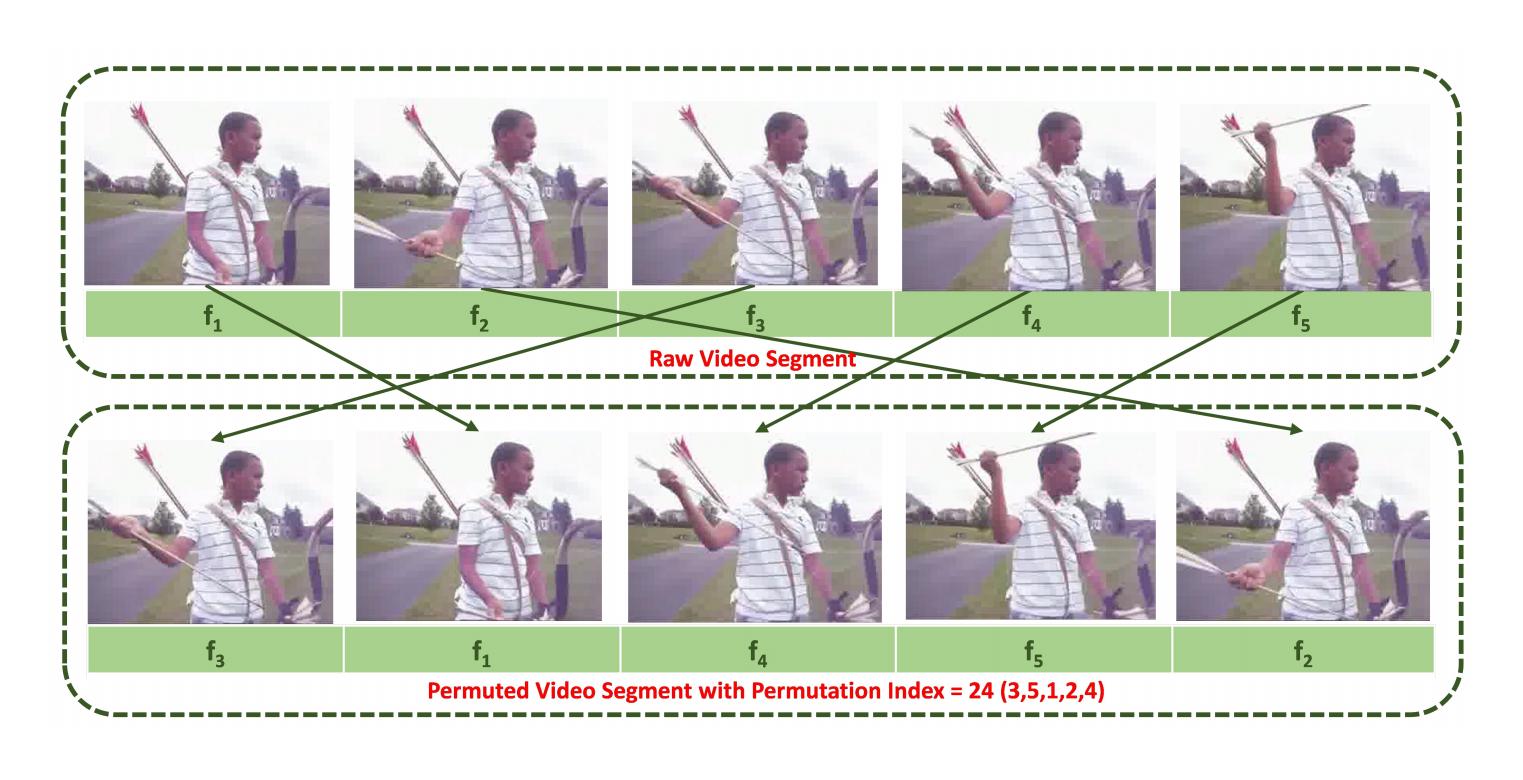}
\end{center}
\caption{{\textbf {Frame Permutation Prediction Task:}} Given a raw video segment, the frames are permuted according to a permutation index randomly drawn from a predefined set of indexes and the network learns to predict this index}
\label{fig:example2}
\end{figure*}

In order to better learn spatiotemporal features in videos for anomaly detection, we introduce the task of guessing the indices of randomly permuted video frames as shown in Fig.~\ref{fig:example2}. While doing so the network implicitly reasons about the object shape, position, and orientation in time without the use of any optical flow information. Thus, the learned feature representation carries rich semantic or structural information. Training for anomaly detection is done with normal videos only. At test time, the network is unable to predict the permutation of video frames from unseen context. Results show that the proposed method achieves significant improvement in anomaly detection over the best available self-supervised visual-representation learning methods.

Our main contributions are two fold. 1) We provide an overview of one-class classification (anomaly detection) using self-supervised visual representation learning. Such analysis do exists for other downstream tasks but not for anomaly detection. 2) We propose a simple method for self-supervised learning of spatiotemporal features from videos without the use of any extra information. The proposed method outperforms numerous other methods in video anomaly detection, which use spatial, temporal or combined self-supervision.

The rest of the paper is structured as follows. Research
related to our work is discussed in Section 2. In Section 3, we
describe the task of permuting video frames for self-supervised visual representation learning. Experimental results and discussion are given in Section 4, and Section 5 concludes the paper.

\section{Related Work}

\subsubsection{Self-Supervised Representation Learning from Images.}

Numerous methods have been proposed for self-supervised representation learning on images, each exploring a different pretext task. Doersch et al. \cite{doersch2015unsupervised:13} perform self-supervised representation learning by predicting the relative position of image patches. Noroozi and Favaro \cite{noroozi2016unsupervised-25} design a jigsaw puzzle game as pretext task, where the model is trained to place nine shuffled patches back to their original locations. Other autoenocder-based methods try to recover part of the data itself, such as image inpainting \cite{pathak2016context-26}, image colorization \cite{zhang2016colorful-14}, and its improved variant channel prediction \cite{zhang2017split-28}. Noroozi et al. \cite{noroozi2017representation-29} introduce a method for self-supervised representation learning that uses an artificial supervision-signal based on counting visual primitives. Gidaris et al. \cite{gidaris2018unsupervised-30} propose to randomly rotate an image by one of four possible angles and let the model predict the rotation. Caron et al. \cite{caron2018deep-31,caron2019unsupervised-32} use a clustering-based approach to generate pseudo-labels.

\subsubsection{Self-Supervised Representation Learning from Videos.}

 Video datasets contain raw spatiotemporal signals, which can be used for self-supervised representation learning. Vondrick et al. \cite{vondrick2018tracking-34} use video colorization as a self-supervised learning problem. Wang and Gupta \cite{wang2015unsupervised-35} propose a way of self-supervised representation learning by tracking moving objects in videos. Another method includes validating frame order \cite{misra2016shuffle-36}, where the pretext task is to determine whether a sequence of frames from a video is placed in the correct temporal order or not. Wei et al. \cite{wei2018learning-37} show that predicting the arrow of time (whether the video is playing forward or backward) learns useful latent representation. HY Lee et al. \cite{lee2017unsupervised-73} use optical flow magnitude to select frames with high motion magnitude and then use sorting task to learn a feature representation. Xu, Dejing, et al. \cite{xu2019self-74} introduce a clip order prediction task as a self-supervised learning task. However, all these pretext tasks are never used for the downstream task of visual anomaly detection in videos.
 
\subsubsection{Anomaly Detection with Images.}

Many researchers have used reconstruction-based methods for anomaly detection. Xia et al. \cite{xia2015learning-20} use a convolutional autoencoder with a regularizing term that produces a large reconstruction error for out-of-distribution samples. An and Cho \cite{an2015variational-21} use a variational autoencoder to extract an anomaly score based on the reconstruction probability estimated through Monte-Carlo sampling. Zhai et al. \cite{zhai2016deep-22} investigate the usage of deep structured energy-based models for anomaly detection, in particular focusing on two decision criteria: energy score and reconstruction error. Zong et al. \cite{zong2018deep-19} propose to jointly model the encoded features and the reconstruction error in a deep autoencoder for anomaly detection. Generative adversarial network-based anomaly detection method is used by Li et al. \cite{li2018anomaly-23}. Ruff et al. \cite{ruff2018deep-42} revisit classical one-class SVMs \cite{scholkopf2000support-40} with deep representations to improve anomaly detection results on complex data. The maximun softmax-probability of a classifier is used by Hendrycks and Gimpel \cite{hendrycks2016baseline-43} for anomaly detection. Lee et al. \cite{lee2017training-44} develop a training method for neural classification networks for better anomaly detection without losing the original classification accuracy. Golan and El-Yaniv \cite{golan2018deep-15} propose that learning a method to discriminate between different geometric transformations applied to normal images encourages learning of features that are useful for anomaly detection.

\subsubsection{Anomaly Detection with Videos.}

Video anomaly detection has mostly been applied to detect anomalies in crowd behaviour, which is important to avert any casualties. Hand-crafted methods \cite{tung2011goal-46,wu2010chaotic-47} propose novel features. A model is trained to learn normal features and an anomaly is detected by identifying the isolated clusters or outliers. In deep learning-based methods, a network is trained to reconstruct frames of normal instances. This trained network cannot reconstruct the anomalous instances, thus, detects anomalies. Hasan et al. \cite{hasan2016learning-48} model regular frames using a 3-D convolutional autoeocder. \cite{chong2017abnormal-49,luo2017remembering-50} uses Convolutional LSTM Auto-Encoder (ConvLSTM-AE) to model both normal appearance and motion at the same time for crowd anomaly detection. Future frame prediction \cite{mathieu2015deep-52} has also been proposed for anomaly detection \cite{liu2018future-51}. A model is trained to predict future video frames of normal training data using the previous frames. In the testing phase, prediction error is used to declare an anomaly.

\section{Frame Permutation Prediction}

Our goal is to use the raw spatiotemporal signals in videos to learn a feature representation carrying rich semantic and structural meaning for the downstream task of visual anomaly detection in videos which has never been addressed before. We learn this representation by solving a frame permutation prediction task. By solving this complex pretext task without any motion information for normal videos, the network learns both low-level and high-level features such that it is able to identify anomalous videos at test time.

\subsection{Problem Formulation}

Given a video with $N$ frames $\{f_1, \dots, f_N\}$, we divide it into $Z=N-M+1$ sub-sequences, each of $M$ consecutive frames. These sub-sequences or video segments $S=\{s_i\}_{i=1}^Z$ are separated by one frame with an overlap of $M-1$. We then permute the frames of each segment according to some permutation index $y_i \in [1, \dots, M!]$ $\forall i$, called pseudo-label of the frame permutation. For each raw input video segment $s_i$, the permutation results in
\begin{equation}
  \label{eq:permutation}
  S^*=\{{s_i}^*\}_{i=1}^Z=\{g(s_i, y_i)\}_{i=1}^Z,
\end{equation}
where $g(.)$ is the permutation operator.\\

Our aim is to train a neural network $W$, with parameters $\psi$, such that it can predict the permutation of jumbled video frames. In other words, training is performed to obtain optimal $\psi$ such that
\begin{equation}
  \label{eq:selfsupervised}
  \argmin_{\psi} \sum_{i} \ell(W_\psi({s_i}^*),y_i),
\end{equation}
where $\ell$ is a general loss function.

\subsection{$M$-Stream Siamese Convolutional Neural Network}

\begin{figure*}[hbt!]
\centering
\includegraphics[height=10.0cm]{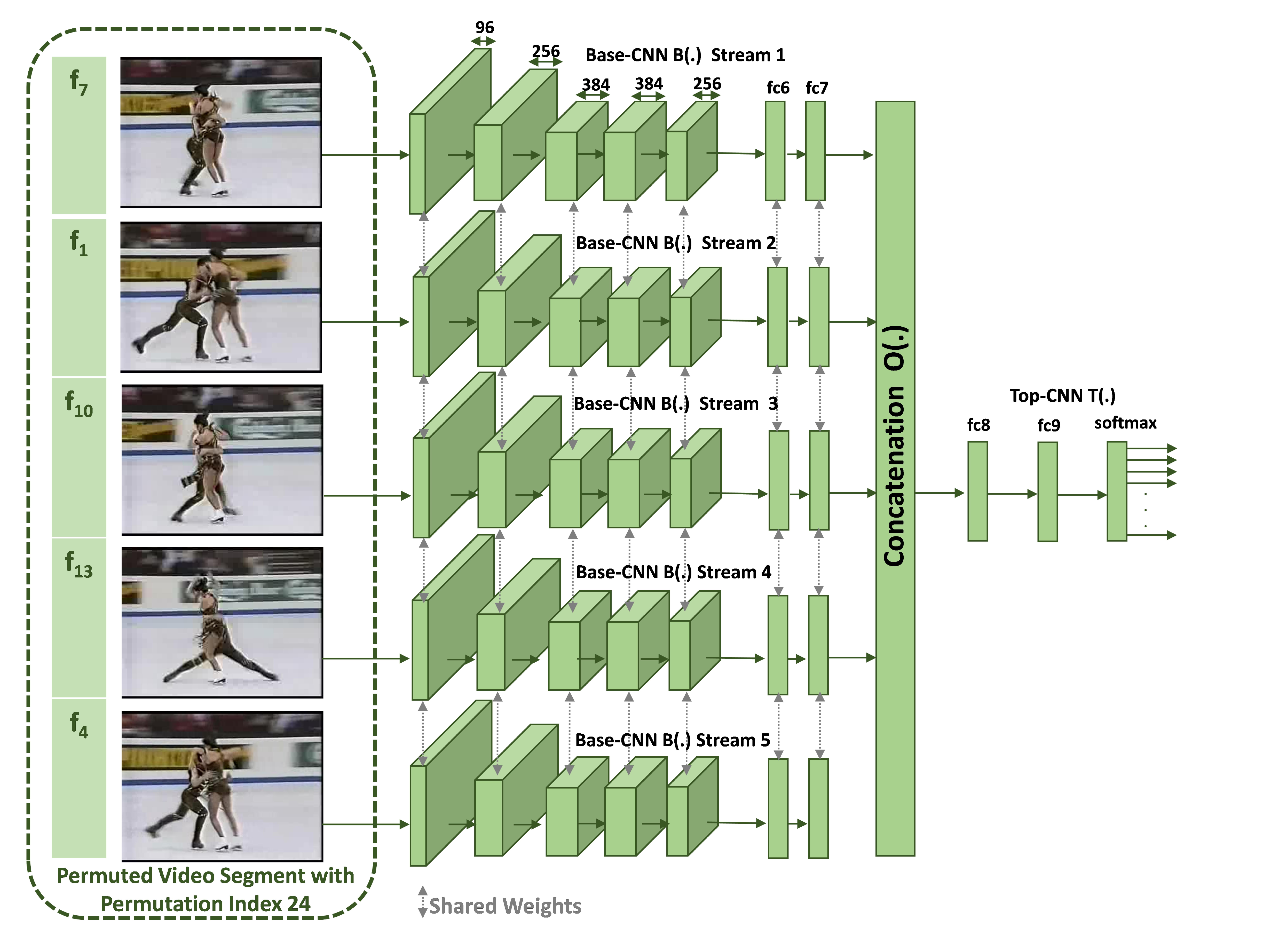}
\caption{{\textbf{5-Stream Siamese CNN:}} Our network architecture has five parallel streams. Each stream from $conv1$ to $fc7$ layer forms the $Base-CNN$ with shared weights. Each stream takes a frame as input, and produces a feature representation. The concatenation operator $O(.)$ concatenates all 5 feature representations, and feeds the output to $Top-CNN$, which produces a conditional probability density function over the 120 possible permutation indexes of the frames. Technically, the output probability should be $1$ at the $24-th$ location and $0$ elsewhere. Further details can be found in the supplementary material}
\label{fig:example4}
\end{figure*}

For this work, we use an $M$-Stream Siamese CNN as shown in Fig.~\ref{fig:example4}, where each stream consists of a Base-CNN (BCNN). All $M$-BCNNs are from \emph{conv1} to \emph{fc7} layer of the CaffeNet architecture \cite{jia2014caffe-72} (1-GPU version of AlexNet \cite{krizhevsky2012imagenet-38}) with shared weights. Each BCNN model B(.) takes one frame $f_k$ of the permuted input video segment ${s_i}^*$, and returns a feature representation $r_k$ of that frame. 
\begin{equation}
\label{eq:BCNN}
  R=\{r_k\}_{k=1}^M = \{B(f_k;\theta)\}_{k=1}^M,
\end{equation}
where $\theta$ are the shared learnable parameters of all BCNN models $B(.)$.\\

The stack of layers from \emph{conv1} to \emph{fc7} making $M$-BCNNs have shared weights, therefore, they have same number of parameters as the AlexNet architecture. All these $M$-representations are concatenated by a concatenation operator O(.) 
\begin{equation}
\label{eq:Concat}
  x_i = O(r_1, r_2, \dots, r_M),
\end{equation}
where $x_i$ is the concatenated feature representation of the frames of the permuted video segment ${s_i}^*$. \\

The layers following the concatenation layer form another CNN called the Top-CNN (TCNN). TCNN model $T(.)$ is a logistic classifier, which takes as input the concatenated representation $x_i$, and yields as output a vector $Y \in \mathbb{R}^{M!}$  with the probability value for each possible label.
\begin{align}
\label{Probability}
  T(x_i)=p(Y|x_i;\phi),
\end{align}
where $p(Y|x_i;\phi)$ is the predicted probability for the permutation labels $Y=\{y_k\}_{k=1}^{M!}$ and $\phi$ represents the learnable parameters of the model $T(.)$\\

\subsection{Training}

Our network is trained to learn better and more detailed information by solving a complex task without the use of any optical flow information. Unlike the previous self-supervised learning approaches our method not only focuses on low-level features but also learns high-level features which make the task of visual anomaly detection such easier and faster. We are not interested in the final performance of the frame permutation prediction task, rather we are only interested in the learned intermediate representation. Notice that during the training on the frame permutation prediction task, we set the stride of the first layer ($conv1$) of our $M$-Stream Siamese CNN to be 2 instead of 4 (CaffeNet model uses 4).

Let the training video be divided into $Z$ permuted segments $S^*=\{{s_i}^*\}_{i=1}^Z$ with corresponding permutations $P=\{y_i\}_{i=1}^Z$. Note that $y_i \in [1, \dots, M!]$ $\forall i$. In order to train the network the parameters of the $M$-Stream Siamese CNN are updated such that
\begin{equation}
\label{eq:theta}
  \theta = \argmin_{\theta}\frac{1}{Z}\sum_{i=1}^Z L(s_i,y_i,\theta,\phi).
\end{equation}
\begin{equation}
\label{eq:phi}
  \phi = \argmin_{\phi}\frac{1}{Z}\sum_{i=1}^Z L(s_i,y_i,\theta,\phi).
\end{equation}
The loss function $L(.)$ is the cross entropy loss defined as

\begin{equation}
\label{eq:loss_fun}
  L(s_i,y_i,\theta,\phi) = - \frac{1}{M!}\sum_{i=1}^{M!}y_i\log(T(O(B(g(s_i, y_i);\theta));\phi)). 
\end{equation}
 $T(O(B(g(s_i, y_i))))$ is the output of the $M$-Stream Siamese CNN. Both $\theta$ and $\phi$ are jointly optimized. 
 
\subsection{Anomaly Detection}

If the training of the pretext task defined in the previous section is done with the normal videos then the network will only be able to predict the permutation labels for them. For out-of-distribution or abnormal examples (videos not seen during training), the predicted probability diverges from the actual label, giving a higher value of the cross-entropy loss compared to the in-distribution examples. 
We normalize this loss function for all video segments in the test data to [0,1] and calculate the anomaly score $A(s_i)$ for each video segment $\{s_i\}_{i=1}^Z$ by the following minmax normalization:
\begin{equation}
\label{eq:anomaly_acore}
  A(s_i) = \frac{L(s_i,y_i,\theta,\phi)-min_i  L(s_i,y_i,\theta,\phi)}{max_i  L(s_i,y_i,\theta,\phi)-min_i L(s_i,y_i,\theta,\phi)}. 
\end{equation}

\section{Experimental Results}
We first show the results of self-supervised learning for anomaly detection over images, and follow it up with videos. We use the best performing method for images with videos as well to identify the role of spatial anomaly detection over videos. The results of our method and other video-based self-supervised methods are also given.

\subsection{Datasets}

We consider four image datasets: CIFAR-10, CIFAR-100 \cite{krizhevsky2009learning-54}, fashion-MNIST \cite{xiao2017fashion-63} and ImageNet \cite{deng2009imagenet-57}, and 2 video datasets: UCF101 \cite{soomro2012ucf101-60} and ILSVRC2015 \cite{russakovsky2015imagenet-61} in our experiments.\\

    {\bf \textbullet \ \ CIFAR-10}
    consists of 69,000 32$\times$32 colored images in 10 classes with 6,000 images per class. There are 50,000 training images and 10,000 test images, equally divided over the classes.
    
    {\bf \textbullet \ \ CIFAR-100}
    is similar to CIFAR-10 but with 100 classes containing 600 images for each class. This set has a fixed train/test partition with 500 training images and 100 test images per class. The 100 classes of this dataset are grouped into 20 superclasses, which we use in our experiments.
    
    {\bf \textbullet \ \ Fashion-MNIST}
    is a dataset comprising 28$\times$28 grayscale images of 70,000 fashion products from 10 categories with 7000 images per category. There are 60,000 training and 10,000 testing images. 
    
    {\bf \textbullet \ \ ImageNet}
    consists of 1000 classes with more than 14 million images. We have used a subset of 20 classes for our anomaly detection experiments. We resize all the images to 256$\times$256.
    
    {\bf \textbullet \ \ UCF101}
    is an action recognition dataset of 13320 videos with 101 action categories. These 101 actions can be divided into 5 types: 1) human-object interaction, 2) body-motion only, 3) human-human interaction, 4) playing musical instruments, and 5) sports. We have used a subset of this dataset with 20 different action classes that includes actions from all the five types. There are 50 training and 10 testing videos per class. Training videos are taken from the train split-1 while testing videos are taken from test split-1 of the UCF101 dataset. 
    
    {\bf \textbullet \ \ ILSVRC2015}
    is an ImageNet VID dataset for object detection in videos. The final release of VID dataset consists of three splits. The training set contains 3862 video snippets with 56 to 458 videos per category. The validation set contains 555 snippets. The test set contains 937 snippets. We have used a subset of this dataset, which consists of videos of 20 different object classes. There are 100 training and 15 testing videos per class. To create the dataset, the testing videos are taken from test or validation set whereas the training videos are taken from the train set.\\
    
    Further details on the dataset can be found in the supplementary material.
    

\subsection{Experimental Setup}

We have used one-vs-all evaluation scheme in our experiments. Suppose a dataset has \emph{C} classes and we train the self-supervised visual learning task on one class \emph{c}, which we call as normal class. Samples from remaining \emph{C-1} classes are said to be abnormal samples. We quantify the performance using the area under the ROC curve metric, which is commonly utilized as a performance measure for anomaly detection models. The same experiment is repeated \emph{C} times, where each time a different normal class is used to train the model. Results are averaged for all \emph{c} normal classes.

\subsection{Anomaly Detection on Images}

We have used solving jigsaw puzzle game (Jigsaw) \cite{noroozi2016unsupervised-25}, image colorization (Color) \cite{zhang2016colorful-14}, image inpainting (Inpaint) \cite{pathak2016context-26}, counting visual primitives (Count) \cite{noroozi2017representation-29}, split brain autoencoders (Split-Brain) \cite{zhang2017split-28}, and predicting image rotations (RotNet) \cite{gidaris2018unsupervised-30} as self-supervised visual representation learning methods \cite{gidaris2018unsupervised-30} for the task of visual anomaly detection. With these simple tasks and no semantic labels, we can learn a powerful visual representation using a CNN, which can be used for the well-known problem of novelty detection. We also include the results of DeepSVDD \cite{ruff2018deep-42}, Geometric Transform (GT) \cite{golan2018deep-15}, and InceptionCAE NN-QED \cite{sarafijanovic2019fast-62} for CIFAR10, CIFAR-100, and fashion-MNIST. The results of these three methods are taken from \cite{golan2018deep-15} and \cite{sarafijanovic2019fast-62} as the evaluation protocol is same as ours.

We trained each of the model for 100 epochs on the normal class. Batch size for all methods was set to 64. The results for CIFAR-10 and CIFAR-100 are given in the Table~\ref{table:cifar10/100}. Interestingly, all the self-supervised techniques perform better than the methods specifically designed for one-class classification (anomaly detection). The visual features learned by self-supervised methods are even better than those learned by supervised or unsupervised methods. This makes the task of visual anomaly detection much easier, faster, and accurate. 
In case of fashion-MNIST, anomaly detection by solving jigsaw puzzle game performs better than RotNet. However, the average AUCROC of deepSVDD is slightly higher than the jigsaw network as given in Table~\ref{table:fashion-mnist}. 

\begin{table*}[h]
\begin{center}
\caption{AUROC of anomaly detection on CIFAR-10 and CIFAR-100. The best performing method in each experiment is in bold. All values are percentages}
\label{table:cifar10/100}
\begin{tabular}{c c c c c c c c c c c c c}
\toprule[1.5pt]
\multirow{ 3}{*}{Dataset} & $c_i$ & Deep & GT & Inception & Count & Color & Inpaint & Split- & Jigsaw & RotNet \\

& & SVDD & & CAE & & & & Brain \\
& & \cite{ruff2018deep-42}. &  \cite{golan2018deep-15} & \cite{sarafijanovic2019fast-62} & \cite{noroozi2017representation-29} & \cite{zhang2016colorful-14} & \cite{pathak2016context-26} &  \cite{zhang2017split-28} & \cite{noroozi2016unsupervised-25} & \cite{gidaris2018unsupervised-30}\\

\midrule
\multirow{11}{*}{\shortstack{CIFAR-10\\($32\times32\times3$)}} 
& 0 & 61.7 & 74.7 & 66.7 & 74.1 & 71.1 & 72.0 & 65.2 & 77.3 & \textbf{80.3} \\
& 1 & 65.9 & 95.7 & 71.3 & 96.1 & 88.4 & 88.9 & 93.6 & \textbf{97.0} & 96.8 \\
& 2 & 50.8 & 78.1 & 66.8 & 79.6 & 70.3 & 71.4 & 78.6 & 79.2 & \textbf{80.0} \\
& 3 & 59.1 & 72.4 & 64.1 & 71.1 & 69.6 & 70.8 & 73.1 & 73.5 & \textbf{75.6} \\
& 4 & 60.9 & 87.8 & 72.3 & 87.4 & 65.2 & 64.3 & 87.6 & 88.6 & \textbf{89.1}\\
& 5 & 65.7 & 87.8 & 65.3 & 88.5 & 66.8 & 70.2 & 86.9 & 89.6 & \textbf{90.1}\\
& 6 & 67.7 & 83.4 & 76.4 & 85.9 & 79.9 & 80.2 & 81.1 & 86.8 & \textbf{87.2}\\
& 7 & 67.3 & 95.5 & 63.7 & 95.9 & 70.5 & 75.6 & 93.4 & 94.8 & \textbf{96.0}\\
& 8 & 75.9 & 93.3 & 76.9 & 94.0 & 68.3 & 84.6 & 83.1 & 93.4 & \textbf{95.7}\\
& 9 & 73.1 & 91.3 & 72.5 & 92.1 & 78.1 & 83.4 & 93.1 & 92.9 & \textbf{93.4}\\
\cdashline{2-11}
& $avg$  & 64.8 & 86.0 & 69.6 & 86.5 & 73.4 & 76.1 & 83.6 & 87.3 & \textbf{88.4}\\
\midrule
\multirow{21}{*}{\shortstack{CIFAR-100\\($32\times32\times3$)}} 
& 0 & 57.4 & 74.7 & 66.0 & 79.2 & 70.1 & 75.4 & 81.6 & 80.4 & \textbf{82.8} \\
& 1 & 63.0 & 68.5 & 60.1 & 71.1 & 54.2 & 58.4 & 61.3 & 73.3 & \textbf{75.2} \\
& 2 & 70.0 & 74.0 & 59.2 & 75.3 & 68.7 & 71.2 & 74.9 & 75.6 & \textbf{77.4} \\
& 3 & 55.8 & 81.0 & 58.7 & 81.6 & 65.3 & 74.6 & 69.0 & 80.2 & \textbf{85.6} \\
& 4 & 69.0 & 78.4 & 60.9 & 76.4 & 62.1 & 65.3 & 68.9 & 78.9 & \textbf{80.1} \\
& 5 & 51.0 & 59.1 & 54.2 & 66.5 & 51.1 & 50.6 & 52.3 & 64.3 & \textbf{67.4} \\
& 6 & 59.9 & 81.8 & 63.7 & 82.9 & 75.4 & 78.7 & 83.6 & 84.2 & \textbf{87.1} \\
& 7 & 53.0 & 65.0 & 66.1 & 66.4 & 61.9 & 63.3 & 65.1 & \textbf{68.2} & 66.3 \\
& 8 & 51.6 & 85.5 & 74.8 & 87.5 & 75.5 & 81.2 & 87.8 & 86.3 & \textbf{89.4} \\
& 9 & 72.9 & 90.6 & 78.3 & 86.9 & 72.1 & 79.9 & 85.4 & 89.1 & \textbf{90.8} \\
& 10 & 81.5 & 87.6 & 80.4 & 86.2 & 68.3 & 72.5 & 75.6 & 88.2 & \textbf{88.3} \\
& 11 & 53.6 & 83.9 & 68.3 & 81.1 & 74.2 & 78.4 & 82.9 & 84.6 & \textbf{85.2} \\
& 12 & 50.6 & 83.2 & 75.6 & 77.5 & 66.5 & 74.6 & 76.4 & 79.2 & \textbf{80.1} \\
& 13 & 44.0 & 58.0 & \textbf{61.0} & 56.3 & 53.2 & 56.2 & 55.2 & 58.1 & 60.3 \\
& 14 & 57.2 & 92.1 & 64.3 & 90.7 & 78.4 & 89.1 & 93.8 & 92.9 & \textbf{94.9} \\
& 15 & 47.7 & 68.3 & 66.3 & 69.9 & 62.1 & 65.8 & 66.2 & 70.4 & \textbf{73.6} \\
& 16 & 54.3 & 73.5 & 72.0 & 73.2 & 57.8 & 62.9 & 65.3 & 74.8 & \textbf{76.4} \\
& 17 & 74.7 & 93.8 & 75.9 & 96.3 & 70.4 & 65.4 & 60.2 & 96.0 & \textbf{97.8} \\
& 18 & 52.1 & 90.7 & 67.4 & 89.4 & 71.1 & 78.1 & 89.6 & 91.5 & \textbf{92.1} \\
& 19 & 57.9 & 85.0 & 65.8 & 85.7 & 76.2 & 80.9 & 85.4 & 86.3 & \textbf{90.6} \\
\cdashline{2-11}
& $avg$  & 58.9 & 78.7 & 67.0 & 79.0 & 66.7 & 71.1 & 74.1 & 80.1 & \textbf{82.1} \\

\bottomrule[1.5pt]
\end{tabular}
\end{center}
\end{table*}

\begin{table*}[ht!]
\begin{center}
\caption{AUROC of anomaly detection on fashion-MNIST dataset. The best performing method in each experiment is in bold. All values are percentages}
\label{table:fashion-mnist}
\begin{tabular}{c c c c c c c c c}
\toprule[1.5pt]

\multirow{ 3}{*}{Dataset} & $c_i$ & Deep & GT & Inception & Jigsaw & RotNet \\
& & SVDD & & CAE \\
& & \cite{ruff2018deep-42} &  \cite{golan2018deep-15} & \cite{sarafijanovic2019fast-62}  & \cite{noroozi2016unsupervised-25} & \cite{gidaris2018unsupervised-30}\\

\midrule
\multirow{11}{*}{\shortstack{Fashion-MNIST\\($28\times28\times2$)}} 
& 0 & 98.8 & \textbf{99.4} & 92.4 & 98.4 & 97.4 \\
& 1 & \textbf{99.7} & 97.6 & 98.8 & 97.9 & 96.6 \\
& 2 & \textbf{93.5} & 91.1 & 90.0 & 91.0 & 92.7 \\
& 3 & 94.9 & 89.9 & \textbf{95.0} & 90.7 & 90.4 \\
& 4 & \textbf{95.1} & 92.1 & 92.0 & 90.9 & 87.9 \\
& 5 & 90.4 & \textbf{93.4} & \textbf{93.4} & 88.4 & 92.0 \\
& 6 & \textbf{98.0} & 83.3 & 85.5 & 97.2 & \textbf{98.0} \\
& 7 & 96.0 & \textbf{98.9} & 98.6 & 96.5 & 98.2 \\
& 8 & 95.4 & 90.8 & 95.1 & 95.7 & \textbf{98.4} \\
& 9 & 97.6 & \textbf{99.2} & 97.7 & 91.2 & 89.5 \\
\cdashline{2-7}
& $avg$  & \textbf{95.9} & 93.5 & 93.9 & 94.5 & 94.1 \\

\bottomrule[1.5pt]
\end{tabular}
\end{center}
\end{table*}

For ImageNet dataset, we compare the results of anomaly detection using different self-supervised visual representation learning methods. Table~\ref{table:imagenet} shows that the RotNet model gives the highest average AUROC for anomaly detection. This is because in order to predict the rotation of an image the network must also learn to model object shape, i.e., it should be able to localize salient objects in the image and recognize their orientation and object type. Learning of low-level object features like color, texture, etc. alone are not enough to predict the image rotations. Therefore, unlike the other self-supervised representation learning methods that mainly focus on low-level features, the RotNet model focuses on learning both low-level and high-level object characteristics, which can better assist in visual anomaly detection. However, for rotationally symmetric objects, the RotNet model will fail to predict rotations, hence will not be able to learn a good feature representation. 

\begin{table*}[h]
\begin{center}
\caption{AUROC of anomaly detection on ImageNet dataset. The best performing method in each experiment is in bold. All values are percentages}
\label{table:imagenet}
\begin{tabular}{c c c c c c c c}
\toprule[1.5pt]

\multirow{ 3}{*}{Dataset} & $c_i$ & Count & Color & Inpaint & Split- & Jigsaw & RotNet \\

& & & & & Brain \\
& & \cite{noroozi2017representation-29} & \cite{zhang2016colorful-14} & \cite{pathak2016context-26} &  \cite{zhang2017split-28} & \cite{noroozi2016unsupervised-25} & \cite{gidaris2018unsupervised-30}\\

\midrule
\multirow{21}{*}{\shortstack{ImageNet\\($256\times256\times3$)}} 
& 0 & 77.6 & 67.6 & 71.2 & 75.4 & 81.5 & \textbf{95.6} \\
& 1 & 83.2 & 74.5 &	76.4 & 80.2 & 89.3 & \textbf{98.6} \\
& 2 & 66.4 & 60.2 & 62.1 & 63.2 & 68.9 & \textbf{76.0} \\
& 3 & 73.1 & 65.5 &	68.7 & 70.1 & 75.9 & \textbf{93.9} \\
& 4 & 67.6 & 63.4 &	64.5 & 62.3 & \textbf{70.7} & 61.8 \\
& 5 & 75.4 & 68.7 &	71.1 & 72.7 & 77.2 & \textbf{77.9} \\
& 6 & 73.2 & 66.1 & 67.6 & 69.9 & 78.4 & \textbf{93.2} \\
& 7 & 74.1 & 65.4 &	68.2 & 70.8 & 76.3 & \textbf{84.4} \\
& 8 & 66.6 & 61.8 &	61.5 & 63.5 & 68.7 & \textbf{86.2} \\
& 9 & 72.9 & 67.6 &	69.8 & 71.2	& 75.2 & \textbf{89.3} \\
& 10 & 84.7 & 75.4 & 78.7 & 81.6 & 87.1 & \textbf{94.7} \\
& 11 & 65.2 & 60.2 & 62.7 & 64.5 & 66.9 & \textbf{83.9} \\
& 12 & 72.6	& 63.8 & 65.5 & 67.8 & 75.0 & \textbf{89.7} \\
& 13 & 73.4	& 66.1 & 68.3 &	70.8 & 70.1 & \textbf{75.9} \\
& 14 & 85.7 & 70.6 & 78.9 &	82.1 & 89.5 & \textbf{94.0} \\
& 15 & 76.8 & 64.3 & 67.6 &	71.6 & 79.6 & \textbf{91.2} \\
& 16 & 68.4 & 63.8 & 66.7 & 65.4 & 71.3 & \textbf{72.6} \\
& 17 & 75.4 & 69.8 & 72.1 &	74.3 & \textbf{77.1} & 76.9 \\
& 18 & 67.5 & 60.2 & 63.4 &	65.9 & 70.8 & \textbf{73.7} \\
& 19 & 82.4 & 71.3 & 76.7 &	78.1 & 85.1 & \textbf{94.9}\\
\cdashline{2-8}
& $avg$ & 74.11 & 66.315 & 69.085 &	71.07 &	76.73 &	\textbf{85.22} \\

\bottomrule[1.5pt]
\end{tabular}
\end{center}
\end{table*}

\begin{table*}[h]
\begin{center}
\caption{AUROC of anomaly detection using state-of-the-art self-supervised visual representation learning baselines and our approach on UCF101 and ILSVRC2015 datasets. AUROC values are an average of 20 AUROCs corresponding to 20 different models trained on exactly one of the 20 classes. Each model's in-distribution examples are from one of the 20 classes, and the test out-of-distribution samples are from the remaining 19 classes. All values are percentages}
\label{table:videos}
\begin{tabular}{c c c c c c c c}
\toprule[1.5pt]
\multirow{3}{*}{Dataset} & Video & Tracking & Shuffle and & AoT & RotNet & Sorting & \textbf{Ours} \\
& Colorization & & Learn \\

& \cite{vondrick2018tracking-34} & \cite{wang2015unsupervised-35} & \cite{misra2016shuffle-36} & \cite{wei2018learning-37} & \cite{gidaris2018unsupervised-30} &
\cite{lee2017unsupervised-73}\\

\midrule
\multirow{1}{*}{\shortstack{UCF101}} 
& 66.3 & 56.4 & 67.5 & 54.1 & 72.8 & 74.6 & \textbf{76.4}\\

\midrule
\multirow{1}{*}{\shortstack{ILSVRC2015}} 
& 69.2 & 70.1 & 70.7 & 61.0 & 74.4 & 73.8 & \textbf{75.5}\\

\bottomrule[1.5pt]
\end{tabular}
\end{center}
\end{table*}

\subsection{Anomaly Detection on Videos}

We have compared the anomaly detection results of our proposed approach (Frame Permutation Prediction Task) with other self-supervised representation learning methods on videos: video colorization \cite{vondrick2018tracking-34}, visual learning by tracking moving objects in videos (Tracking) \cite{wang2015unsupervised-35}, shuffle and learn \cite{misra2016shuffle-36}, AoT \cite{wei2018learning-37}, and sorting \cite{lee2017unsupervised-73}. Overall best performing image-based self-supervised representation learning method, i.e., RotNet is also used for video anomaly detection. This is done to see the importance of spatial features in video anomaly detection. With RotNet, we treat the video frames of UCF101 and ILSRVC2015 datasets as independant images, and perform visual anomaly detection similar to the image datasets.  

For video anomaly detection using frame permutation prediction task, the video segments are generated from both normal and abnormal class videos. A label zero is assigned to a video segment if it belongs to normal class and one if it belongs to abnormal class. We use stochastic gradient descent optimizer with a momentum of 0.9. The training uses colored frames of normal video segments, which are resized to 225$\times$225 pixels. The batch size is kept 10 for all anomaly detection experiments. Training is done for 100 epochs with a learning rate of $10^{-3}$. 

The comparison is shown in Table~\ref{table:videos}. The spatiotemporal features learned by our proposed frame permutation prediction task are better than the prior methods. Unlike the previous methods our network not only captures low-level video features but also focuses on high-level spatiotemporal features that make the task of visual anomaly detection much more easier and faster. It is also observed that the temporal features alone provide less meaningful information than the spatial features for the task of visual anomaly detection in videos. This is the reason AOT gives poor video anomaly detection results. This suggests that learning good spatial features is more important than learning good temporal features for visual anomaly detection. However, it can be seen from our experimental results that learning both spatiotemporal features from normal videos allows better detection of anomalous videos than learning spatial features alone provided that the the learned representation carries rich semantic and structure understanding of the video content. The results in Table~\ref{table:videos} show that our model better learns the semantic and structural meanings of the normal class (in-distribution examples), therefore, making it easier to detect the abnormal class (out-of-distribution examples). The average AUROC obtained by our method exceeds the existing methods both on UCF101 and ILSVRC2015 datasets.

It is also found that the number of frames, $M$, in a video segment and the corresponding possible permutations, $M!$, is an important hyperparameter. Empirical results in Table~\ref{table:different M} show that $M=5$ gives optimal results. Fewer frames and the corresponding fewer permutations do not provide enough information to learn, as the frames in a segment would be quite similar. Larger values of $M$ do not improve the performance.

\begin{table}[h]
\begin{center}
\caption{AUROC of anomaly detection with different $M$}
\label{table:different M}
\begin{tabular}{c c c c c}
\toprule[1.5pt]
\multirow{2}{*}{Dataset} & \multirow{2}{*}{$M=3$} & \multirow{2}{*}{$M=4$} & \multirow{2}{*}{$M=5$} & \multirow{2}{*}{$M=6$} \\ 
\\
\midrule
\multirow{1}{*}{\shortstack{UCF101}} 
& 69.4 & 72.4 & \textbf{76.4} & 76.1\\

\midrule
\multirow{1}{*}{\shortstack{ILSVRC2015}} 
& 70.5 & 73.5 & \textbf{75.5} & 75.3\\

\bottomrule[1.5pt]
\end{tabular}
\end{center}
\end{table}

Experiments in Table~\ref{table:different combinations} also show that for each video segment instead of choosing a combination of five consecutive frames $\{f_1, f_2, f_3, f_4, f_5\}$, selecting a combination by skipping two frames $\{f_1, f_4, f_7, f_{10}, f_{13}\}$ gives optimal results. This is because consecutive frames tend to be very similar, not allowing the network to learn enough spatiotemporal information.

\begin{table}[h]
\begin{center}
\caption{AUROC of anomaly detection with different skip}
\label{table:different combinations}
\begin{tabular}{c c c c c}
\toprule[1.5pt]
\multirow{2}{*}{Dataset} &
\multirow{2}{*}{$Skip=0$} & \multirow{2}{*}{$Skip=1$} &
\multirow{2}{*}{$Skip=2$} &
\multirow{2}{*}{$Skip=3$}\\
\\
\midrule
\multirow{1}{*}{\shortstack{UCF101}} 
& 67.4 & 73.1 & \textbf{76.4} & 70.2\\

\midrule
\multirow{1}{*}{\shortstack{ILSVRC2015}} 
& 72.5 & 73.3 & \textbf{75.5} & 71.6\\

\bottomrule[1.5pt]
\end{tabular}
\end{center}
\end{table}

The results in Table~\ref{table:cifar10/100}, \ref{table:fashion-mnist}, \ref{table:imagenet}, and \ref{table:videos} clearly show that image-based self-supervised representation learning shows competitive performance. However, there is much room for improvement in self-supervised methods for visual anomaly anomaly detection in videos which has never been addressed in past.

\section{Conclusion}

In this paper, we explore anomaly detection in images and videos using self-supervised visual representation learning. A number of pretext tasks have been proposed fro representation learning but they have never been evaluated for the downstream task of visual anomaly detection. We performed an extensive experimental comparison of anomaly detection in images using existing self-supervised visual representation methods and state-of-the-art algorithms specifically designed for anomaly detection. The best performing image-based self-supervised representation learning method is then used for video anomaly detection to see the importance of spatial features. For visual anomaly detection in videos, we introduce an efficient frame permutation prediction task which learns better spatiotemporal features without the use of any additional information. The proposed method results in improved visual anomaly detection on two video datasets as compared to the state-of-the-art self-supervised representation learning methods.

%
\IEEEpeerreviewmaketitle

\bibliographystyle{IEEEtran}
\bibliography{IEEEabrv,new}

\end{document}